\documentclass[a4paper, 10pt, onecolumn, draftcls, conference]{ieeeconf}      

\IEEEoverridecommandlockouts                              

\usepackage{verbatim}
\usepackage{amsmath}
\usepackage{graphicx}
\usepackage{amssymb}
\usepackage{array}
\usepackage[draft]{fixme}
\usepackage[english]{babel}
\usepackage{verbatim}
\usepackage{subfigure}

\usepackage{booktabs, tabularx}
\usepackage{multirow}

\usepackage{color}

\newcommand{\frm}[1]{\langle #1\rangle}

\DeclareMathOperator{\sig}{sig}

\newcommand{\FriW}[1]{\emph{FriWalk}}

\newcommand{\gas}[2][]{%
	\ifthenelse{\equal{#1}{}}{%
		\textcolor{blue}{(Gas: #2)}%
}{%
		\textcolor{blue}{#1 (Gas: #2)}%
	}%
}

\title{\LARGE \bf Generating Reliable and Efficient Predictions of
  Human Motion: \\A Promising Encounter between Physics and Neural
  Networks} \author{Alessandro Antonucci, Gastone Pietro Rosati
  Papini, Luigi Palopoli, Daniele Fontanelli 
}

\begin{document}

\maketitle
\thispagestyle{empty}
\pagestyle{empty}


\begin{abstract}
  Generating accurate and efficient predictions for the motion of the
  humans present in the scene is key to the development of effective
  motion planning algorithms for robots moving in promiscuous areas,
  where wrong planning decisions could generate safety hazard or
  simply make the presence of the robot ``socially'' unacceptable.
  Our approach to predict human motion is based on a neural network of
  a peculiar kind. Contrary to conventional deep neural networks, our
  network embeds in its structure the popular Social Force Model, a
  dynamic equation describing the motion in physical terms. This
  choice allows us to concentrate the learning phase in the aspects,
  which are really unknown (i.e., the model's parameters) and to keep
  the structure of the network simple and manageable. As a result, we
  are able to obtain a good prediction accuracy with a small
  synthetically generated training set, and the accuracy remains
  acceptable even when the network is applied in scenarios quite
  different from those for which it was trained. Finally, the choices
  of the network are ``explainable'', as they can be interpreted in
  physical terms. Comparative and experimental results prove the
  effectiveness of the proposed approach.
\end{abstract}


\section{Introduction}
\label{sec:introduction}

When a robot travels across a human populated area, the paths it takes
have to be in accordance with the motion of the by-standers. Thereby,
the robot motion planner has to rely on an accurate prediction of how
the humans are going to move in a time horizon of a few seconds.  The
input usable to this purpose can be of various kind. In a foreseeable
future, the robot could use the facial expression and the pose of the
different parts of a person's body as predictors of her motion intent.
But, perceiving and interpreting the body language is currently beyond
the reach of the technology, at least for commonly used low cost
service robots.  Much easier to treat is the story of a person's
positions in the near past, which can be used to predict her motion
assuming a certain degree of temporal continuity in her choices. Many
physics based models can be used to this purpose, the most famous
being the Social Force Model (SFM)~\cite{helbing1995social}. In the
SFM a person is seen as a particle acted on by attractive forces (the
goals) and repulsive forces (the obstacles). The model has known
limitations; one of the most important is that modelling a person as a
particle does not differentiate between motion patterns that are
``natural'' and others that are possible but not frequently taken
(e.g., sideways motions).  These issues can be addressed by leveraging
a relatively high sampling rate and/or by integrating the preferential
nonholonomic behaviour of the human motion into the
model~\cite{farina2017walking,arechavaleta2008nonholonomic}.

An important problem to tackle in order to use the SFM is how to
estimate its many parameters and in particular the intensity and the
direction of the attractive and of the repulsive forces that animate
the motion.  A first possibility is to make heuristic
``rule-of-the-thumb'' choices, but this option is workable only in
very specific conditions, e.g., interaction between a robot and a human
in free space~\cite{col13a,bevilacqua2018reactive}.

Neural Networks (NN) hold the promise to solve this type of problems
in a simpler way. In principle, a deep neural network (DNN) trained
with a sufficient number of samples could learn the human motion
patterns by discovering the underlying dynamic model on its
own. However, the number of layers and of neurons required to manage
the complexity of human behaviours can be very large and is anyway
hard to predict.  Equally difficult is to understand the number of
samples that are needed to train a network of this
complexity. Finally, the use of a DNN lacks a property of remarkable
importance for many applications: the so-called
``explainability''. When an autonomous system takes a decision it is
important to understand why that specific choice has been made, in
order to solve bugs or attribute legal
responsibilities~\cite{pasquale2017toward}. The total absence of a
prior model in a DNN makes explainability hard or even impossible to
achieve.

In this paper, we seek to bridge the gap between model based and
learning based approaches in order to retain the advantages of
both. Our goal is to predict the motion of a human for several seconds
ahead using a short segment of past observations. To this end, we use
a NN, but the network's structure is chosen so that its connections
reflect the dynamics of the SFM, i.e. we embed our prior knowledge
into the NN in the form of a model assuming that the latter acceptably
represents the dynamics of human motion. This way, the learning phase
is concentrated on the aspects for which we actually lack any real
knowledge: the parameters and the forces acting in the SFM. The
advantage of this approach are manifold: 1. wiring a model inside the
NN reduces the number of neurons by a significant amount (we estimate
one or two orders of magnitude), 2. as shown in our experiments, a
relatively small number of synthetically generated samples is
sufficient to generate accurate predictions, even for scenarios that
are quite different from the ones considered in the training set,
3. because our NN retains the model inside, its decisions can be
explained in physical terms, which simplifies the interpretation of
the results of the NN and the explanation of its possible mistakes.

The paper is organised as follows. In Section~\ref{sec:background}, we
report about the related work in the area and we summarise some
background knowledge on the SFM, which will prove useful in the
development of the paper. In Section~\ref{sec:nn}, we report the key
contribution of the paper: how to embed the SFM into the structure of
a NN. In Section~\ref{sec:experiments}, we report a full set of
experiments proving the validity of the approach. Finally, in
Section~\ref{sec:conclusion} we state our conclusions and announce
future work directions.



\section {Background Material}
\label{sec:background}

\subsection{Related Work}
\label{subsec:related}

Physics-based methods for human motion prediction are based on the
explicit dynamical model of Newton’s laws of motion. Their
implementation is quite easy and they usually work faster than
alternative approaches. A major limitation is that they work well in
describing local dynamics, but fail to follow the medium/long-term
intentions of human beings. Estimating on-the-fly the hypothetical
target of walking pedestrian from the past motion is still an open
issue~\cite{rudenko2019human}. For example, in~\cite{luber2010people},
a virtual goal is chosen as the position that a person would reach if
s/he moved with constant velocity, while in~\cite{ikeda2013modeling} a
set of trajectory sub-goals are estimated from the recorded data in a
structured environment. Furthermore, \cite{kretz2018some} proposed a
modified formulation of the SFM to calibrate the parameters with
observable features from empirical data.  

These limitations persist even if NNs are employed. As in this paper,
other works have explored the feasibility of combining the SFM with
machine learning techniques. For instance, a gradient descent based
method was proposed by~\cite{wan2017learning} to learn the parameters
of the interaction force of the SFM; \cite{zhang2020direction}
combined the SFM with three different direction-decision predictors,
namely a Linear Regression, a Neural Network and a Decision Tree
model, and also investigated the environment features that affect the
direction choice.  Conversely, \cite{johansson2007specification} used
an evolutionary learning algorithm to optimise the SFM parameters to
video recorded data of a crowd.  However, neural networks usually
suffer from overfitting, or they depend strictly on the type of
information with which they are trained. For example,
\cite{scholler2020constant} showed that state-of-the-art neural models
can be outperformed by a simple constant velocity model in the case of
linear trajectories.

The idea of pre-wire the network structure with the
physical law for human motion prediction is rather new. A similar
approach has been followed for vehicle dynamics modelling
in~\cite{daLio2019modelling}, where the flexibility of the data-driven
approaches is combined with a NN embedded with the vehicle physics
model. The advantage to have this kind of network is leading the
learning, decreasing the number of trainable parameters, avoiding
overfitting, and reducing the number of examples for the training.

\subsection{Social force model}
\label{subsec:sfm}

In the SFM~\cite{helbing1995social}, the $i$-th pedestrian with mass
$m_i$ and radius $r_i$ in position
$\mathbf{p}_i = \left[x_i, y_i\right]^{T}$ expressed in the frame
$\frm{F} = \{X_f, Y_f\}$ moves towards his/her target at a certain
desired walking speed $v^d_i$ and following a second order dynamic. At
the same time, the motion is perturbed by the environment, e.g. fixed
obstacles, walls, furnitures, etc., and other agents in the
environments. Omitting the subscript $i$ for readability, the total
force $\mathbf{f}$ that acts on the pedestrian is then
$\mathbf{f} = \mathbf{f}^o + \mathbf{f}^e$, i.e.
\begin{equation}
  m \dot{\mathbf{v}} = \mathbf{f}^o + \sum_{j(\neq i)} \mathbf{f}_{j}^p + \sum_{w} \mathbf{f}_w^W ,
	\label{eq:sfm}
\end{equation}
where $\mathbf{v} = \dot{\mathbf{p}}$.  Furthermore, the attractive
force $\mathbf{f}^o$ is defines as
\begin{equation}
	\mathbf{f}^o = m \frac{v^d(t) \mathbf{e}^d(t) - \mathbf{v}(t)}{\tau}
	\label{eq:sfm-attr}
\end{equation}
where the characteristic time $\tau > 0$ parameter determines the rate
of change of the velocity vector, while $\mathbf{e}^d$ is the unit
vector pointing towards the goal. The force exerted by the static
obstacle $w$ on the $i$-th pedestrian is
\begin{equation}
\begin{aligned}
	\mathbf{f}_w^W = & A e^{\left( r-d_w \right) / B} \mathbf{n}_w + k_1 g\left( r-d_w \right) \mathbf{n}_w + \\
	& - k_2 g\left( r-d_w \right) \left(\mathbf{v} \cdot \mathbf{t}_w \right) \mathbf{t}_w,
	\end{aligned}
\label{eq:sfm-rep}
\end{equation}
i.e. the sum of a repulsive component, a compression force and a
sliding friction force.  We denote with
$d_w = ||\mathbf{p} - \mathbf{p}_w||$ the distance between the
pedestrian centre of mass and the coordinates of the obstacle closest
point, so that $\mathbf{n}_w = (\mathbf{p} - \mathbf{p}_w)/d_w$ and
$\mathbf{t}_w = \left[-\mathbf{n}_w(2), \mathbf{n}_w(1)\right]^{T}$
are the distance unit vector and its tangential direction,
respectively. The function $g(x) = \text{max}\{0,x\}$ models the fact
that both the compression and the sliding friction forces exist only
if the pedestrian touches the obstacle ($d_w > r$). $A$, $B$, $k_1$
and $k_2$ are the model parameters. Notice that in this paper we are
neglecting the interaction forces $\mathbf{f}_{j}^p$ with the $j$-th
pedestrian in~\eqref{eq:sfm}, which will be the objective of future
works.

\section{Neural network model}
\label{sec:nn}

In the proposed structured network, the neurons are organised in order
to process the input signals according to~\eqref{eq:sfm}. Two separate
branches, according to the considered scenario, are designed to
estimate the SFM forces. In the case of no obstacles (first scenario),
the agent moves freely towards its goal, so it is subject only
to~\eqref{eq:sfm-attr}. In the second scenario, the pedestrian is
affected by the repulsive force~\eqref{eq:sfm-rep} of the surrounding
static objects (if any). Consequently, each network branch models
effects of different nature, i.e. attractive or repulsive forces,
which, given the linear nature of the SFM, are summed up at the end to
give the resulting force.

\subsection{Open environment}
\label{subsec:nn_net1}

While freely moving towards the desired goal, the pedestrian is only
affected by the force $\mathbf{f}^o$ in~\eqref{eq:sfm-attr}. Hence,
the first neural network \textit{Net1} 
has to predict the two force components $f^o_x$, $f^o_y$.  The network
inputs are $n$ samples of the past $\mathbf{p}$ coordinates of the
pedestrian (up to the current time $t$). In order to avoid spatial
biases, the coordinates are normalised with respect to the first
sample of the window, so that
\begin{equation}
	\Delta\mathbf{p}(t) = \left[\mathbf{p}^{t-n} , \dots, \mathbf{p}^{t}\right] - \mathbf{p}^{t-n}.
	\label{eq:inputs}
\end{equation}
First, two hidden layers with no biases and with only one fully
connected output neuron learn the instantaneous velocity $v_x$, $v_y$
on the $X_f$ and $Y_f$ axis, respectively. Since these two layers are
followed by a $\tanh(\cdot)$ activation function, another single
neuron with no bias is used in each layer to rescale the
estimates. Moreover, the most recent relative motion measurement
$\Delta\mathbf{p}_1(t) = \mathbf{p}^t - \mathbf{p}^{t-1}$ is used to
estimate the components of the normalised goal-directed unit vector
$\mathbf{e}^d$. Finally, the velocities magnitudes derived as
\begin{equation*}
\begin{split}
	\Delta\mathbf{p}'(t) &= \Delta\mathbf{p}(t) - \Delta\mathbf{p}(t-1),
	\\
	\mathbf{D}(t) &= \left[||\Delta\mathbf{p}'(t)_1||, \dots, ||\Delta\mathbf{p}'(t)_{n-1}|| \right]
\end{split}
\end{equation*}
are used to estimate the desired speed $v^d$. All the estimates pass
trough a Lambda layer where they are combined and weighted according
to the $m$ and $\tau$ parameters in~\eqref{eq:sfm-attr}.  The
\textit{Net1} output is then the estimate of $\mathbf{f}^o = \left[f^o_x, f^o_y\right]^{T}$.
The formulation~\eqref{eq:sfm-attr} is translated in the
form of a structured NN as
\begin{equation}
  \mathbf{f}^o = \overbrace{\sig\left( \mathbf{D}(t) W_{v} \right)w_{v s}}^{\frac{m v^d}{\tau}}\overbrace{\frac{\Delta\mathbf{p}(t)_1}{||\Delta\mathbf{p}(t)_1||}}^{\mathbf{e}^d(t)} - \overbrace{\tanh\left( \Delta\mathbf{p}(t) W_\mathbf{v} \right)\odot w_{\mathbf{v}s}}^{\frac{\mathbf{v}(t) m}{\tau}},
  \label{eq:net1}
\end{equation}
where $W_{v}\in \mathbb{R}^{10\times n-1}$, $w_{v s}\in\mathbb{R}^{10}$, $W_{\mathbf{v}}\in\mathbb{R}^{2\times20}$, $w_{\mathbf{v} s}\in\mathbb{R}^{1 \times2}$,
are the weight matrices.
The $\sig(\cdot)$ sigmoid activation function is used
to keep positive the $v^d$ estimate, which then are rescaled by the
weight matrix $w_{v s}$.  The number of learnable
parameters for \textit{Net1} is $123$, where $100$ of them are
entirely devoted to the desired velocity estimates.

\subsection{Structured environment}
\label{subsec:nn_net2}

For the environment with obstacles, the second network \textit{Net2}
has two parallel branches that predict the $f^o_x$, $f^o_y$ and
$f^W_x$, $f^W_y$ components of the force,
respectively. 
The presence of obstacles makes the prediction of the agent's target
indeterminate, i.e. a motion observation cannot be exploited to
understand if the travelled path depends on the obstacle or on the
attraction of the goal.  For this reason, we directly provide
$\mathbf{e}^d$ as input to the branch that estimates the attractive
force $\mathbf{f}^o$. Our strategy for choosing the goal position
(and, hence, $\mathbf{e}^d$) is described later in
Section~\ref{subsec:multigoal}.  The force due to static obstacles is
described by~(\ref{eq:sfm-rep}). In order to simplify the learning
complexity, we neglected the compression and the sliding friction
forces, both in the SFM simulations and in the neural network.  The
second branch of the network then comprises a Lambda layer (followed
by two single neuron layers with no bias) that takes as inputs the
distance $d_w$ and the components of the unit vector $\mathbf{n}_w$ at
the current time $t$. The inputs are combined in an exponential form
as in~(\ref{eq:sfm-rep}), where the only two learnable weights reflect
the $A$, $B$ parameters of the SFM.  The formulation of the total
force $\mathbf{f}$ in the structured NN form is then
\begin{equation}
  \begin{aligned}
    \mathbf{f} = & \sig\left( \mathbf{D}(t) W_{v} \right)w_{v s} \mathbf{e}^d(t) - \tanh\left( \Delta\mathbf{p}(t) W_\mathbf{v} \right)\odot w_{\mathbf{v}s} + \\
    & +\left(w_A e^{ d_w(t)/ w_B} \mathbf{n}_w(t)\right) \odot w_{\mathbf{f}s} ,
    \label{eq:net2}
  \end{aligned}
\end{equation}
where again $w_A, w_B\in\mathbb{R}^{1}$, and
$w_{\mathbf{f}s}\in\mathbb{R}^{1\times2}$ are learning weights.

\subsection{Multi-goal prediction}
\label{subsec:multigoal}

Since the \textit{Net2} network cannot estimate the agent's goal on
its own, we implement a multi-goal approach to estimate the most
likely navigation direction. The proposed strategy is based on the
combination of pseudo-Kalman filtering and likelihood analysis. More
precisely, we make the assumption that the agent never stops walking
in the middle of a corridor (that is, walls are mere obstacles and not
points of interest), so s/he wants to reach one of the possible exits
of the scenario.  To account for corridor sudden stops, the multi-goal
prediction can also model this situation.  Therefore, knowing the
initial measured position, we generate different goal hypotheses and
then generate with the \textit{Net2} a trajectory for each goals.  To
choose one of the possible hypothesised motion trajectories, the
Multiple Model Approach (MMA) presented in~\cite{bar2004estimation} is
used. Specifically, the first-order generalised pseudo-Bayesian
estimator is applied to each predicted trajectory, and the most likely
hypothesis, computed using the likelihood of the generated trajectory
with the observations, is selected.

\section{Experiments}
\label{sec:experiments}


First we prove how the network can be trained using synthetic
datasets. Both {\em Net1} and {\em Net2} were trained with the Adam
optimiser with a learning rate of $0.005$, batch size $128$, and
number of epochs $300$.  First, we created two synthetic sets of
trajectories generated by the SFM in an open (for {\em Net1}) and in a
structured (for {\em Net2}) environments, the latter having two
intersecting corridors.

\begin{figure}[htb!]
  \centering
  \begin{tabular}{cc}
    \includegraphics[width=0.45\columnwidth]{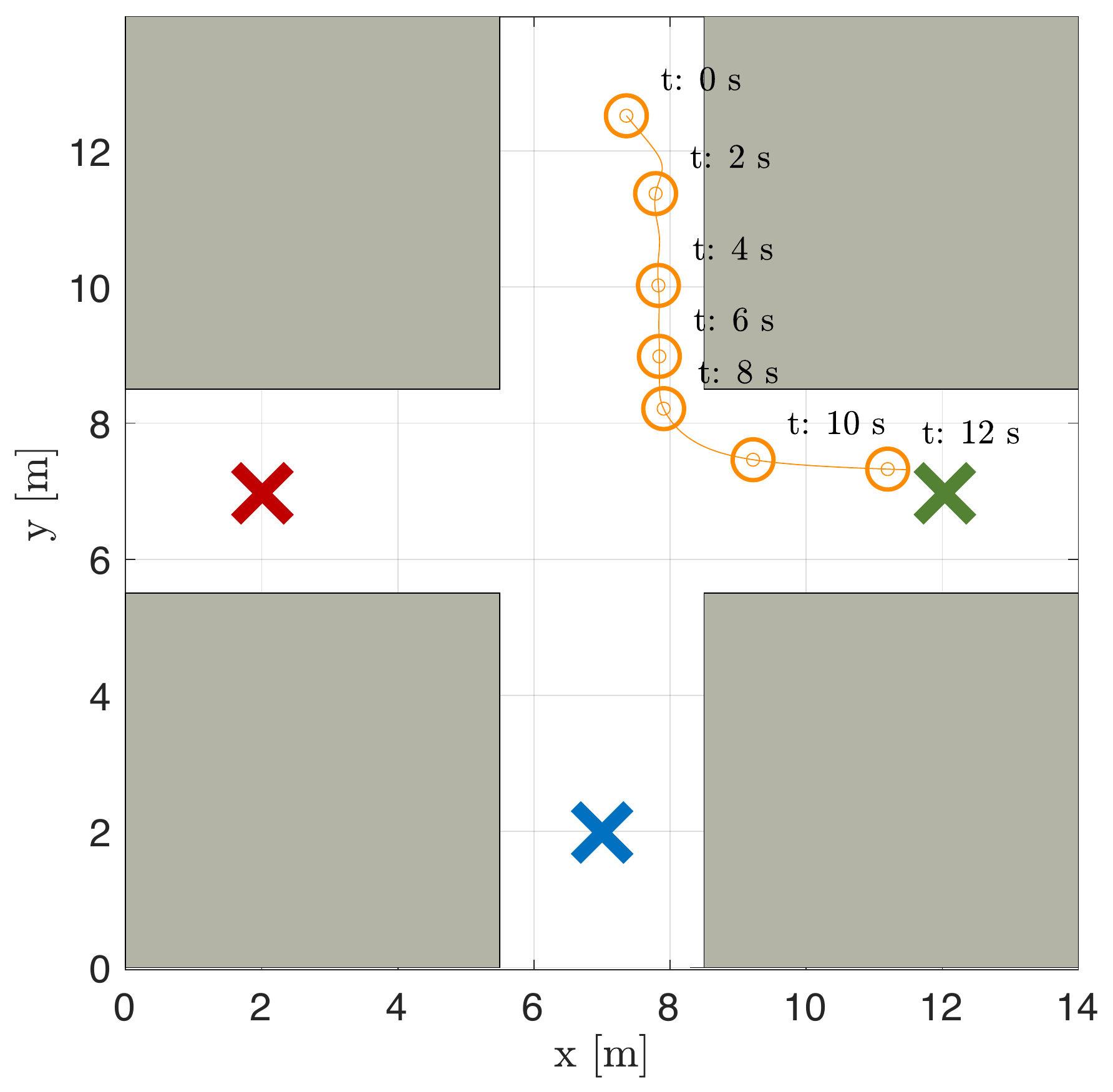} & 
                                                                 \includegraphics[width=0.4\columnwidth]{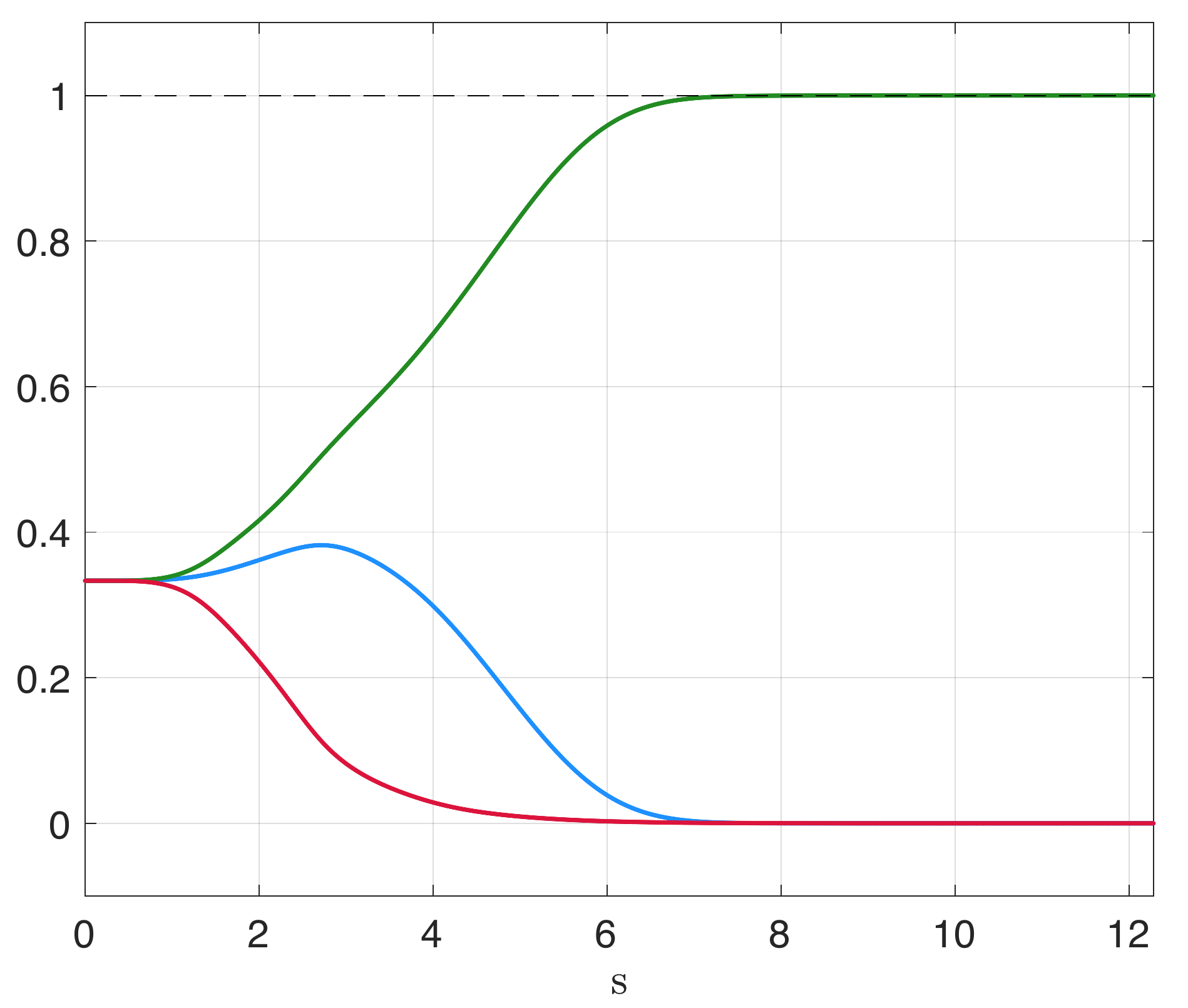} \\
    (a) & (b) \\
  \end{tabular}
  \caption{(a) The agent moves from the uppermost area to the
    \textit{green} exit on the right. The coloured crosses represent
    all the possible waypoints of the scenario. (b) Probabilities of
    the three waypoints.}
  \label{fig:classify}
\end{figure}

In the first set, we run $800$ simulations of $20$~seconds with a
sampling time of $0.1$~s. The initial positions were randomly chosen
within a range between $8$ and $10$~m from the final goal, and, for
each simulation, a random set of parameters were taken from the
intervals $\left[50,90\right]$~kg for $m$, $\left[0.5,0.9\right]$~s
for $\tau$ and $\left[0.5,3\right]$~m/s$^2$ for $v^d$. For the second
set, we run $1200$ simulations where the agent moves through the
corridors intersection, starting from one of the four possible
waypoint areas (see Fig.~\ref{fig:classify}-a) and reaching another
one.
We set the parameters in~\eqref{eq:sfm-rep} to $A = 1000$, $B = 0.08$,
according to~\cite{farina2017walking}.  The $70\%$ of each synthetic
dataset was used as the training sets, while the remaining samples
were used for validation. Notice that, in order to avoid possible
correlations between training and validation, the samples
randomisation is done after dividing the two sets.  The window of the
motion observations was empirically set to $n=10$ samples, which
provides a good trade off between learning speed and network
prediction accuracy. This result is consistent with the fact that the
networks mostly depend on the most recent data, and that a longer
motion observation does not significantly improve the prediction
accuracy~\cite{scholler2020constant}.  Both networks are learned
without over-fitting on their respective datasets, with training mean
squared errors of $8.98 \cdot 10^{-5}$~N and $1.20 \cdot 10^{-3}$~N
for {\em Net1} and {\em Net2}, respectively.

\begin{figure}[htb!]
  \centering
  \begin{tabular}{c}
    \includegraphics[width=0.5\columnwidth]{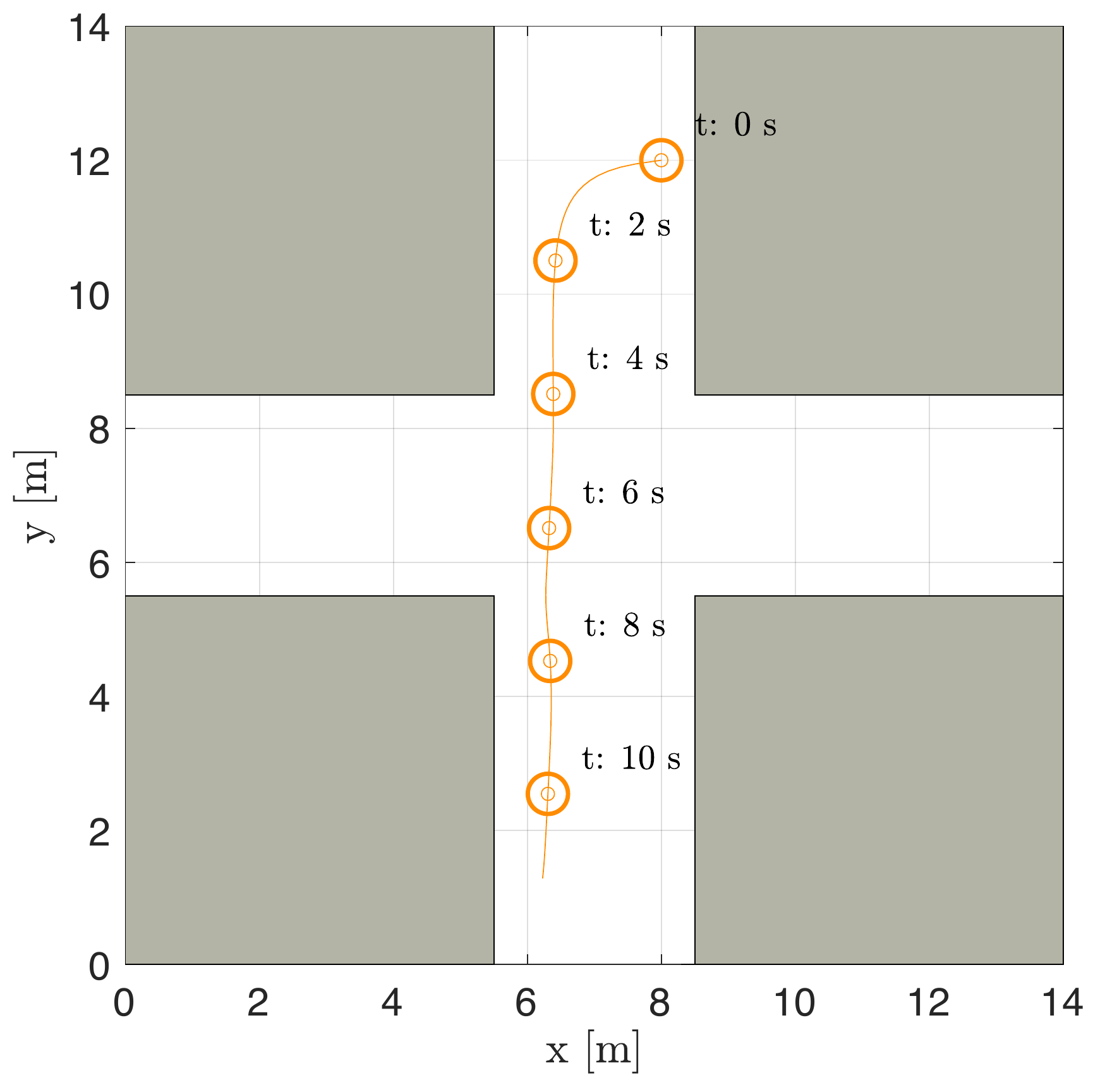} \\
    (a) \\
    \includegraphics[width=0.65\columnwidth]{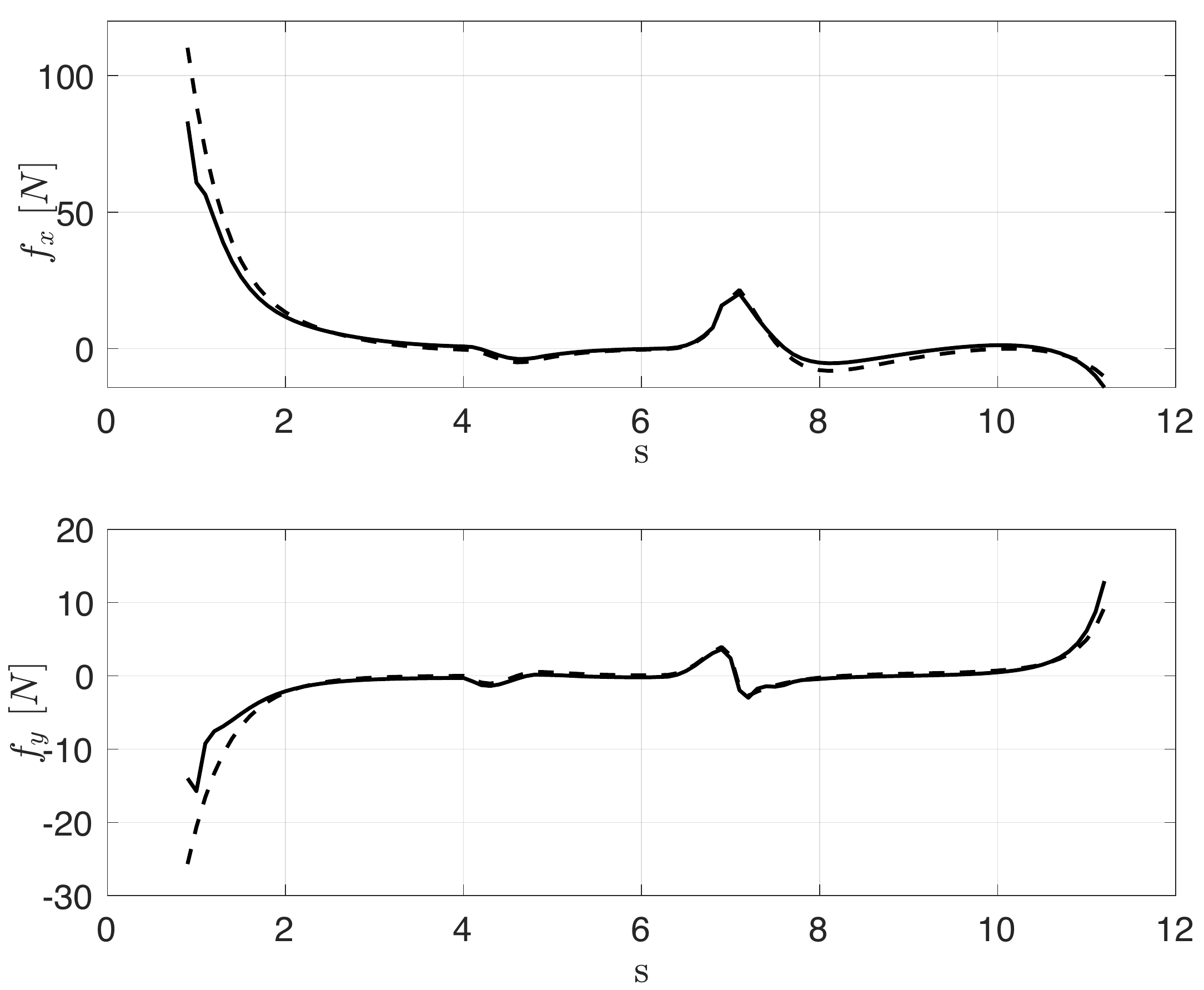} \\
    (b)
  \end{tabular}
  \caption{First validation example. (a) SFM simulated trajectories
    and (b) {\em Net2} forces predictions (solid) compared with the
    SFM forces (dashed).}
  \label{fig:prediction1}
\end{figure}

\begin{figure}[htb!]
  \centering
  \begin{tabular}{c}
    \includegraphics[width=0.5\columnwidth]{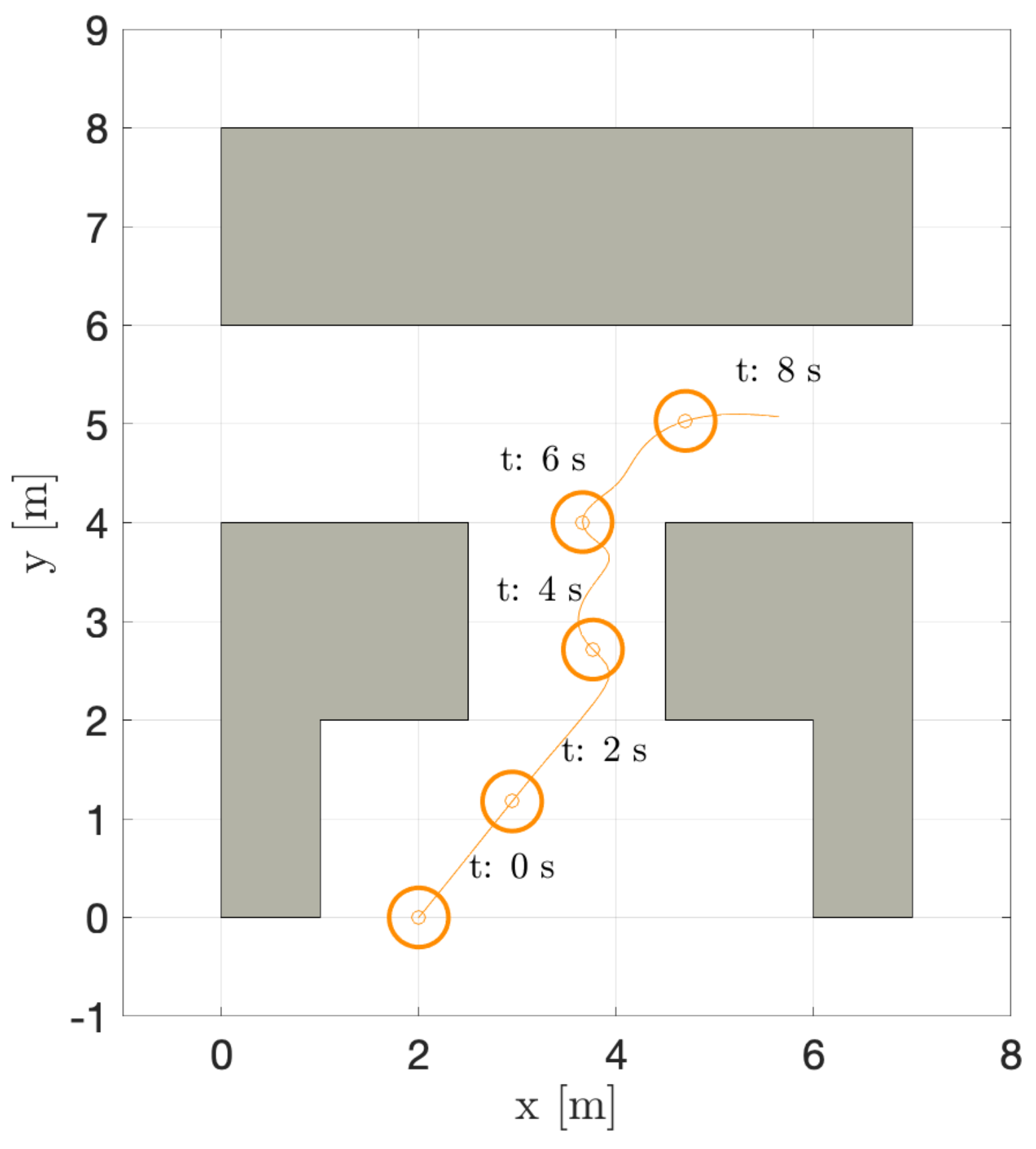} \\
    (a) \\
    \includegraphics[width=0.65\columnwidth]{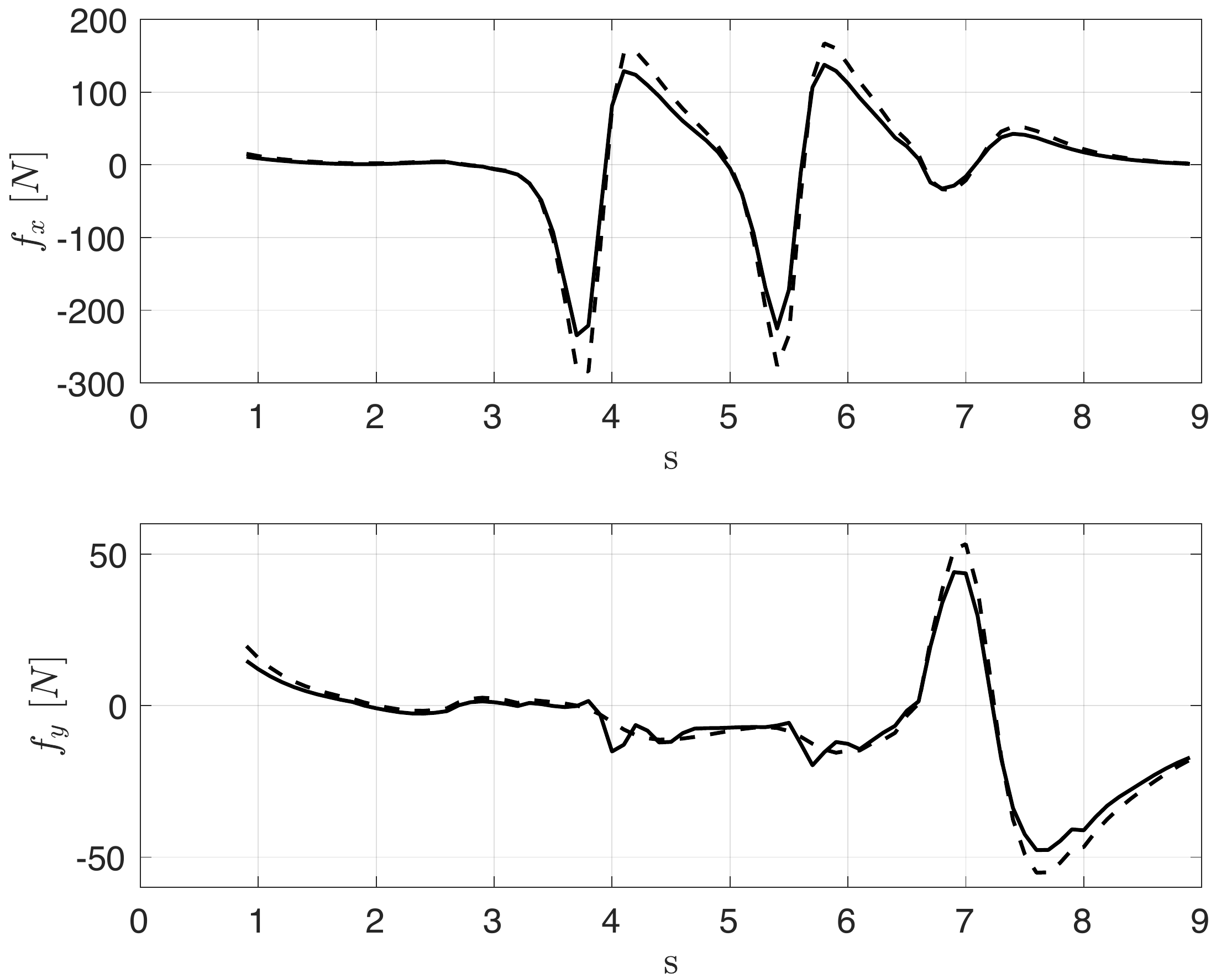} \\
    (b)
  \end{tabular}
  \caption{Second validation example. (a) SFM simulated trajectories
    and (b) {\em Net2} forces predictions (solid) compared with the
    SFM forces (dashed).}
  \label{fig:prediction2}
\end{figure}

\begin{figure}[htb!]
  \centering
  \begin{tabular}{cc}
    \includegraphics[width=0.5\columnwidth]{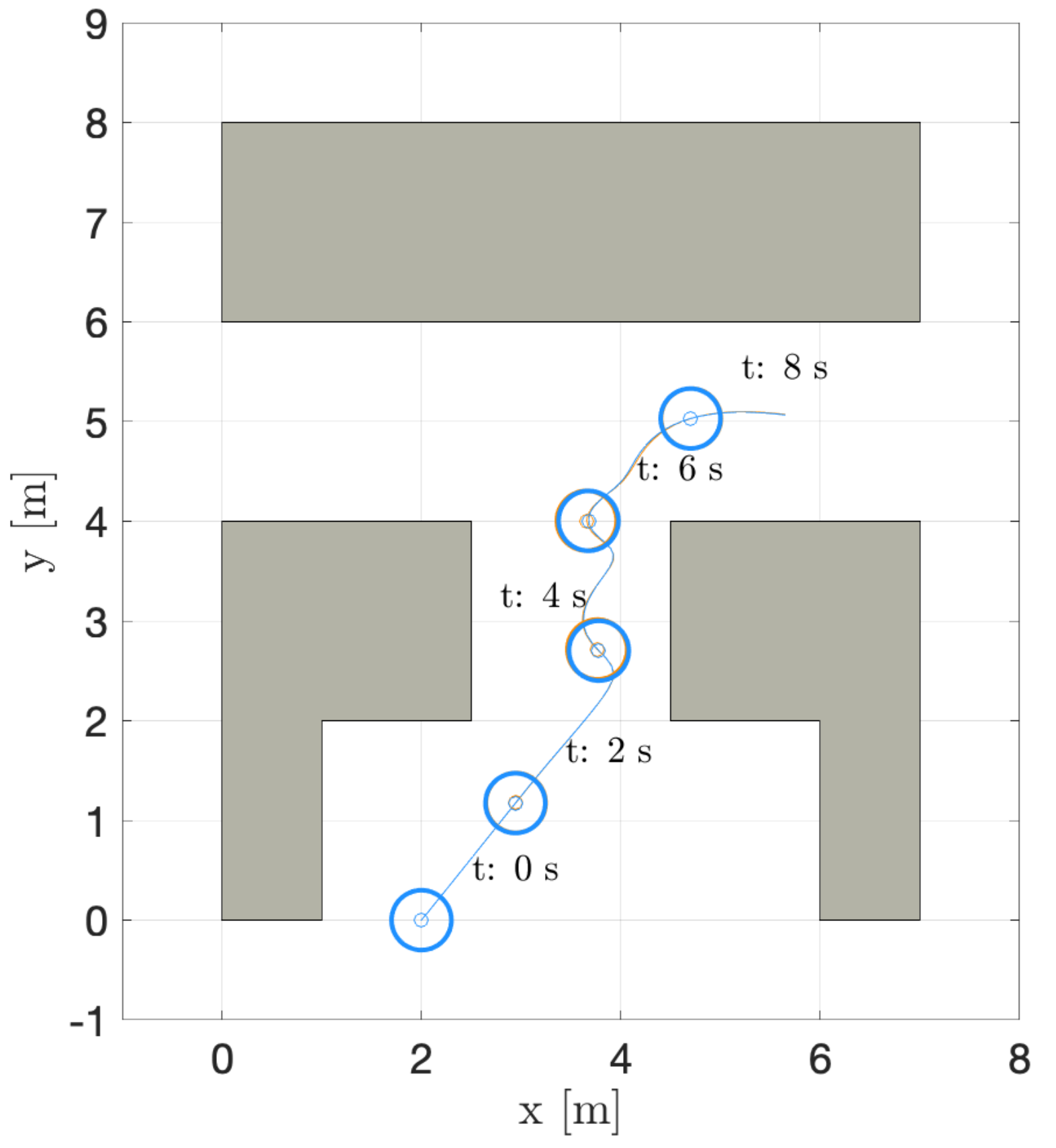} &
                                                                 \includegraphics[width=0.43\columnwidth]{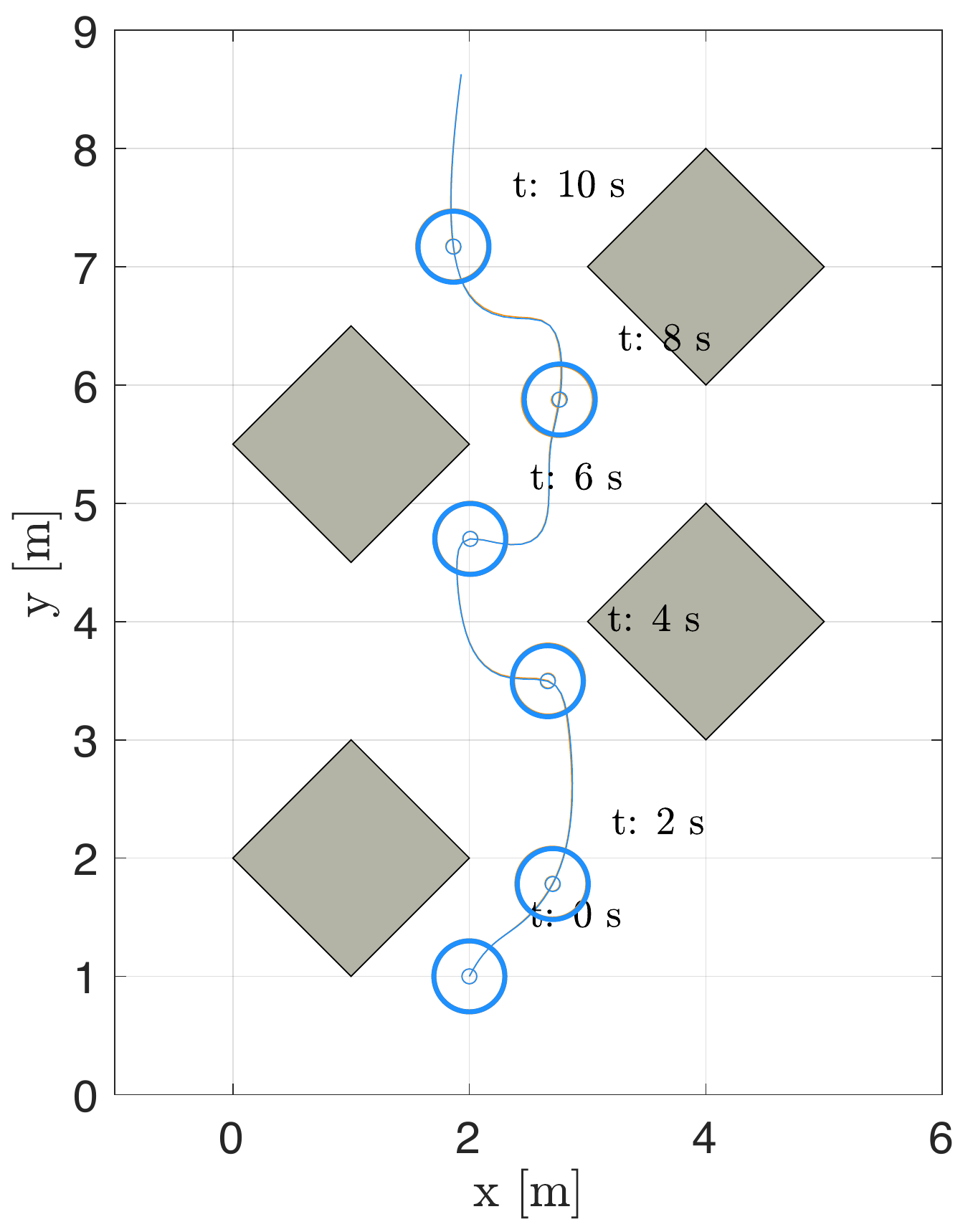} \\	
  \end{tabular}
  \caption{Examples of predicted trajectories using {\em Net2} in a
    structured environment.}
  \label{fig:openloop}
\end{figure}

For space limits, we only report in Fig.~\ref{fig:prediction1} and
Fig.~\ref{fig:prediction2} the results of the force predictions of the
{\em Net2} network (similar results are obtained for {\em Net1}).
For this evaluation, we first generated the entire trajectory, then
passed all the samples to the network and inferred the force
components for each sample. Fig.~\ref{fig:prediction1}-a and
Fig.~\ref{fig:prediction2}-a depict the trajectories generated by the
SFM, while Fig.~\ref{fig:prediction1}-b and
Fig.~\ref{fig:prediction2}-b the comparison between the real and the
network predicted force components $f_x = f_x^o + f_x^W$ and
$f_y = f_y^o + f_y^W$. While Fig.~\ref{fig:prediction1} reports the
validation in the same environment of the synthetic dataset (notice
how the prediction is reliable both in the case of a major repulsive
component around the first second and in the subsequent prevalence of
the attractive component of the force, as reported in
Fig.~\ref{fig:prediction1}-b), Fig.~\ref{fig:prediction2} shows the
results in a completely novel environment, where the agent wants to
reach the exit on the right corridor. Despite a slight underestimation
of the forces in the occurrence of the horizontal collisions with the
walls (see Fig.~\ref{fig:prediction2}-b), the prediction remains
consistent also in this novel scenario. This demonstrate that {\em
  Net2} has not been negatively affected by the environmental biases
during the training.  We then compared the open loop predictions made
by the NN with respect to the trajectories generated with the
SFM. After inferring the forces with the first window of ground truth
motion, we recursively use the predictions using~\eqref{eq:sfm}.
The force inputs used by the network depend solely on the position predicted in the previous time window of $n = 10$ steps, thus generating the results in Fig.~\ref{fig:openloop}.
It can be noticed that in both scenario the trajectory is pretty well replicated for the different scenario reported.

\subsection{{\em Net1} experimental validation}
\label{subsec:validation_net1}

To experimentally validate the performance of the {\em Net1} network
and make a comparison with other methods in the literature, we used
two widely known human motion datasets: the
ETH~\cite{pellegrini2009you} and UCY~\cite{lerner2007crowds} dataset.
The former contains the scene \textit{Hotel} and \textit{ETH}, while
the latter contains three scenes, namely \textit{UCY}, \textit{Zara1}
and \textit{Zara2}. These datasets contain real world pedestrian
trajectories in open scenario, where the influence of the static
obstacles is mostly negligible.

According to other related works~\cite{alahi2016social,
  rudenko2019human, scholler2020constant}, we compute the errors
using: (i) the \textit{Mean Euclidean Distance} (MDE), that is the
average Euclidean distances between all the points of the ground truth
and the predicted trajectory at the same temporal instant; (ii) the
\textit{Final Displacement Error} (FDE), i.e. the distance between the
last predicted position and the last ground truth position.  Unlike
the leave-one-out approach for the cross-validation training used for
example in~\cite{alahi2016social, rudenko2019human}, we used the
synthetically learned {\em Net1} to validate the prediction accuracy
over real-world data. In other works, the observation window was
usually set to $8$ timesteps (that is, $3.2$~s according to the data
acquisition frame rate of the datasets), while the predictions spanned
the successive $4.8$~s. In our model, we observe only $1$~s of the
real-world trajectories, according to the length of the motion
observation window of $n = 10$ samples only, and similarly predict for
$4.8$~s for comparability.
In Table~\ref{table:errors} we show the prediction errors for all the
datasets and the comparison with the Constant Velocity (CV), the
Constant Accelerated (CA) models and Feed Forward (FF) neural network
implemented by~\cite{scholler2020constant}.
\begin{table}[t]
	\centering
	\caption{Prediction errors with our model (SMF-NN) and other state-of-the-art models on the real-world datasets. The metrics are reported in meters.}
	\label{table:errors}
	\begin{tabularx}{0.9\linewidth}{XXXXXXX}
		\toprule
		Metric & Dataset & CV & CA & FF & LSTM & SFM-NN \\
		\midrule
		\multirow{5}{*}{MDE} & {\em Hotel} & 0.27 & 0.95 & 1.59 & 0.15 & 0.36 \\
		& {\em ETH} & 0.58 & 1.35 & 0.67 & 0.60 & 0.68 \\
		& {\em UCY} & 0.46 & 0.79 & 0.69 & 0.52 & 0.46 \\
		& {\em Zara1} & 0.34 & 0.59 & 0.39 & 0.43 & 0.35 \\
		& {\em Zara2} & 0.31 & 0.50 & 0.38 & 0.51 & 0.38 \\
		\midrule
		& Average & 0.39 & 0.84 & 0.74 & 0.44 & 0.45 \\
		\midrule
		\multirow{5}{*}{FDE} & {\em Hotel} & 0.51 & 2.41 & 3.12 & 0.33 & 0.82 \\
		& {\em ETH} & 1.15 & 3.29 & 1.32 & 1.31 & 1.63 \\
		& {\em UCY} & 1.02 & 2.03 & 1.38 & 1.25 & 1.12 \\
		& {\em Zara1} & 0.76 & 1.50 & 0.81 & 0.93 & 0.85 \\
		& {\em Zara2} & 0.69 & 1.30 & 0.77 & 1.09 & 0.95 \\
		\midrule
		& Average & 0.83 & 2.11 & 1.48 & 0.98 & 1.07 \\
		\bottomrule
	\end{tabularx}
\end{table}
Moreover, we report the comparison with the LSTM network
by~\cite{alahi2016social}.  Our model predicts with a very good
accuracy the trajectories in each dataset, with an average MDE of
about $40$~cm with the exception of {\em ETH} scenario. The errors
worsen in the case of the final displacements: this result is mainly
attributable to the strong non-linearity of the real-world
trajectories, which are not easily followed by {\em Net1} in open
loop. Notwithstanding the worsen training conditions of the proposed
approach, our results provides similar (and sometimes better)
performance than the other methods in the literature.

\subsection{Multi-goal prediction on synthetic trajectories}
\label{subsec:multigoal_experiments}

Let us consider again the intersecting corridors scenario. Following
the method described in Section~\ref{subsec:multigoal}, we identify
four waypoints areas, which correspond indeed to the areas chosen for
the network training.  The pedestrian starts from one of the areas
(from the uppermost one in the experiment in
Fig.~\ref{fig:classify}-a), and reaches one of the three
below. Therefore, we generate three different hypotheses for the
trajectory predictions with the {\em Net2} network (one per area,
respectively) choosing as waypoint the area centroids.  As shown in
Fig~\ref{fig:classify}-b, after about $2$~seconds, the classifier is
able to find the correct goal, since our model moves the agent
accordingly with the SFM. In the following seconds the confidence
towards the simulated trajectory increases.

\subsection{{\em Net2} validation}
\label{subsec:validation_net2}

\begin{figure}[htb!]
  \centering
  \begin{tabular}{c}
    \includegraphics[width=\columnwidth]{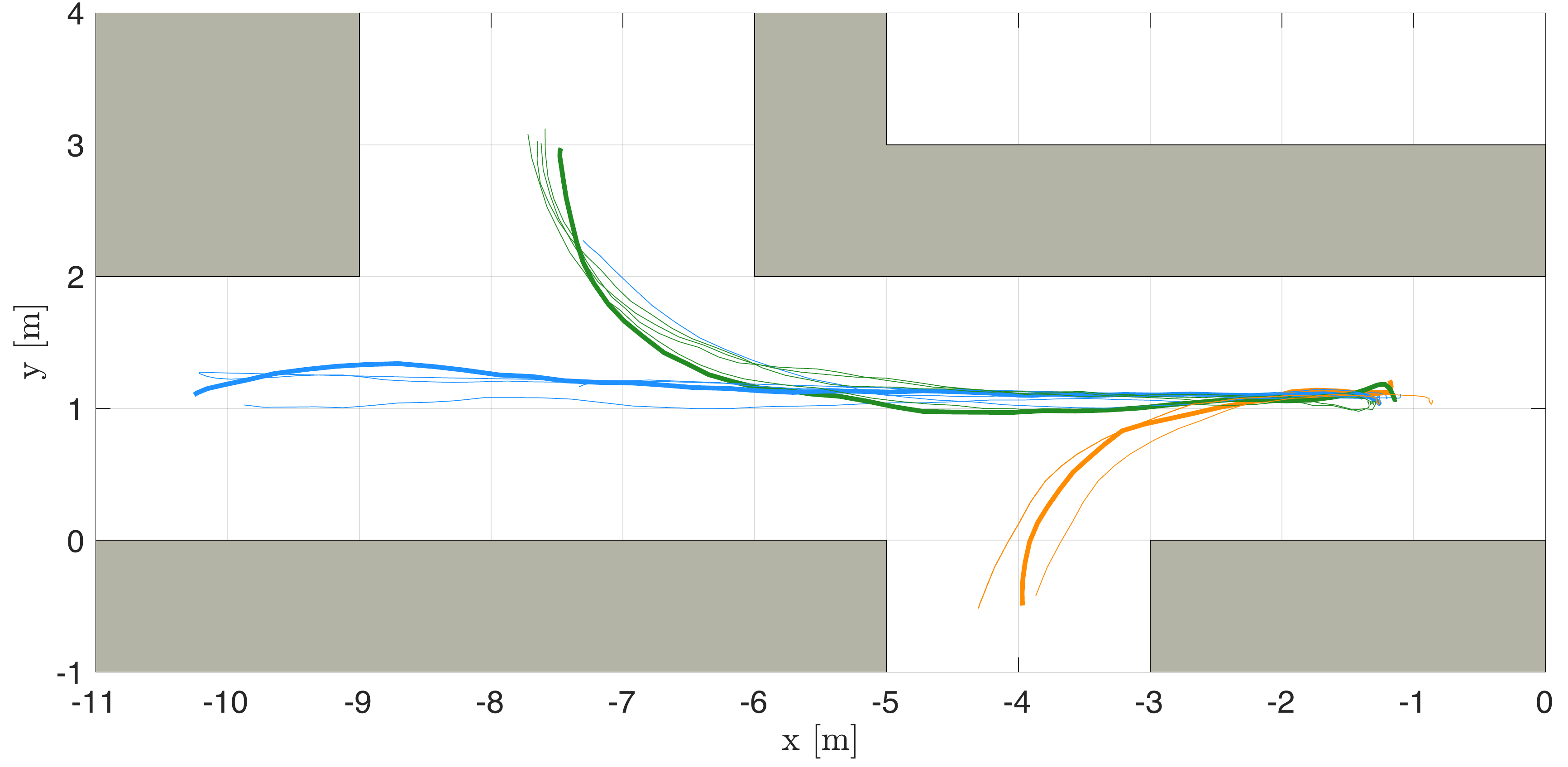} \\
    (a) \\
    \includegraphics[width=0.5\columnwidth]{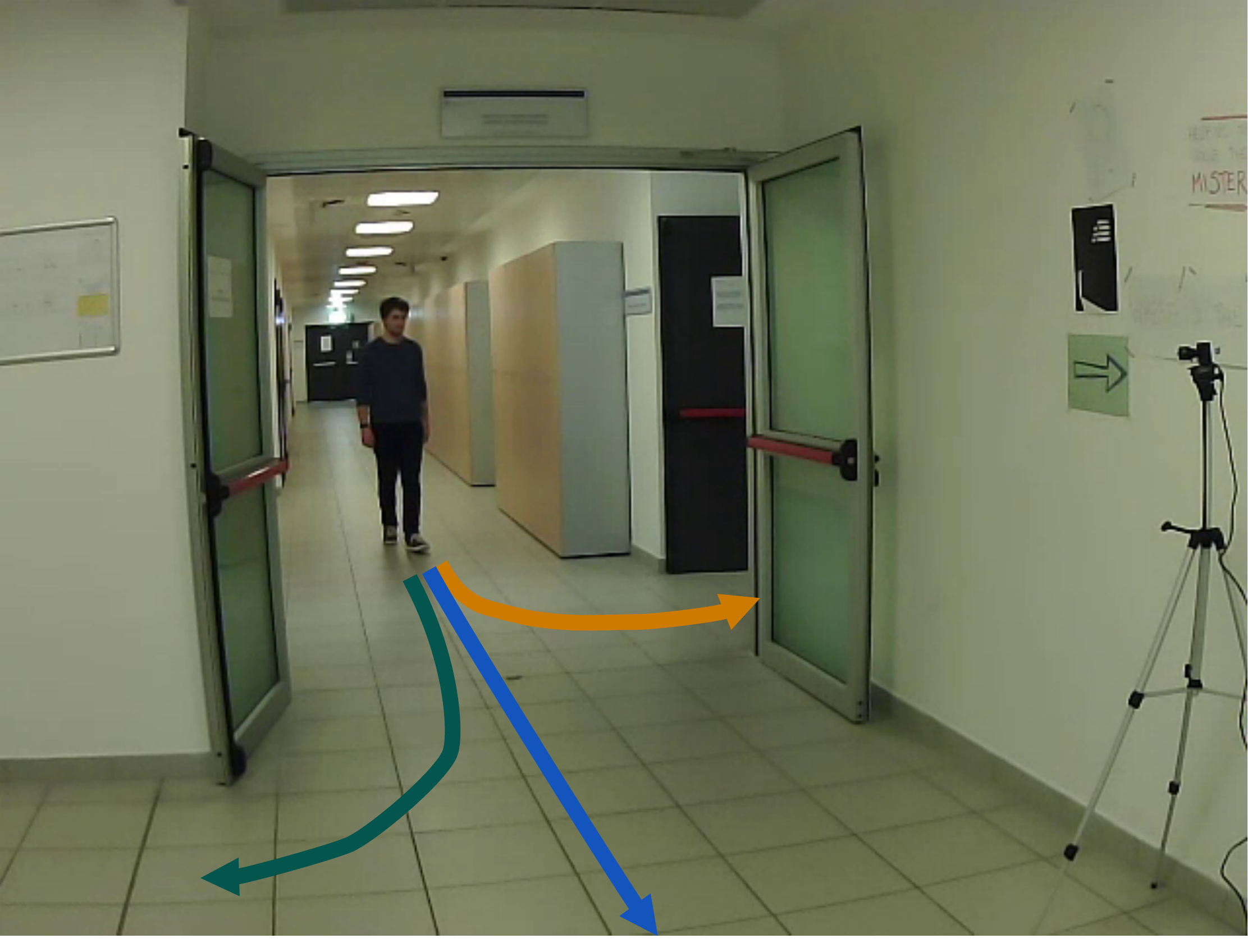} \\
    (b)
  \end{tabular}
  \caption{Experimental trajectories in a hallway. (a) Collected
    trajectories: bold lines are the ones used for the classifier
    evaluation. (b) Experimental set-up: the arrows show the path from
    the starting position towards each destination.}
  \label{fig:povo}
\end{figure}

\begin{figure}[htb!]
  \centering
  \begin{tabular}{c}
    \includegraphics[width=\columnwidth]{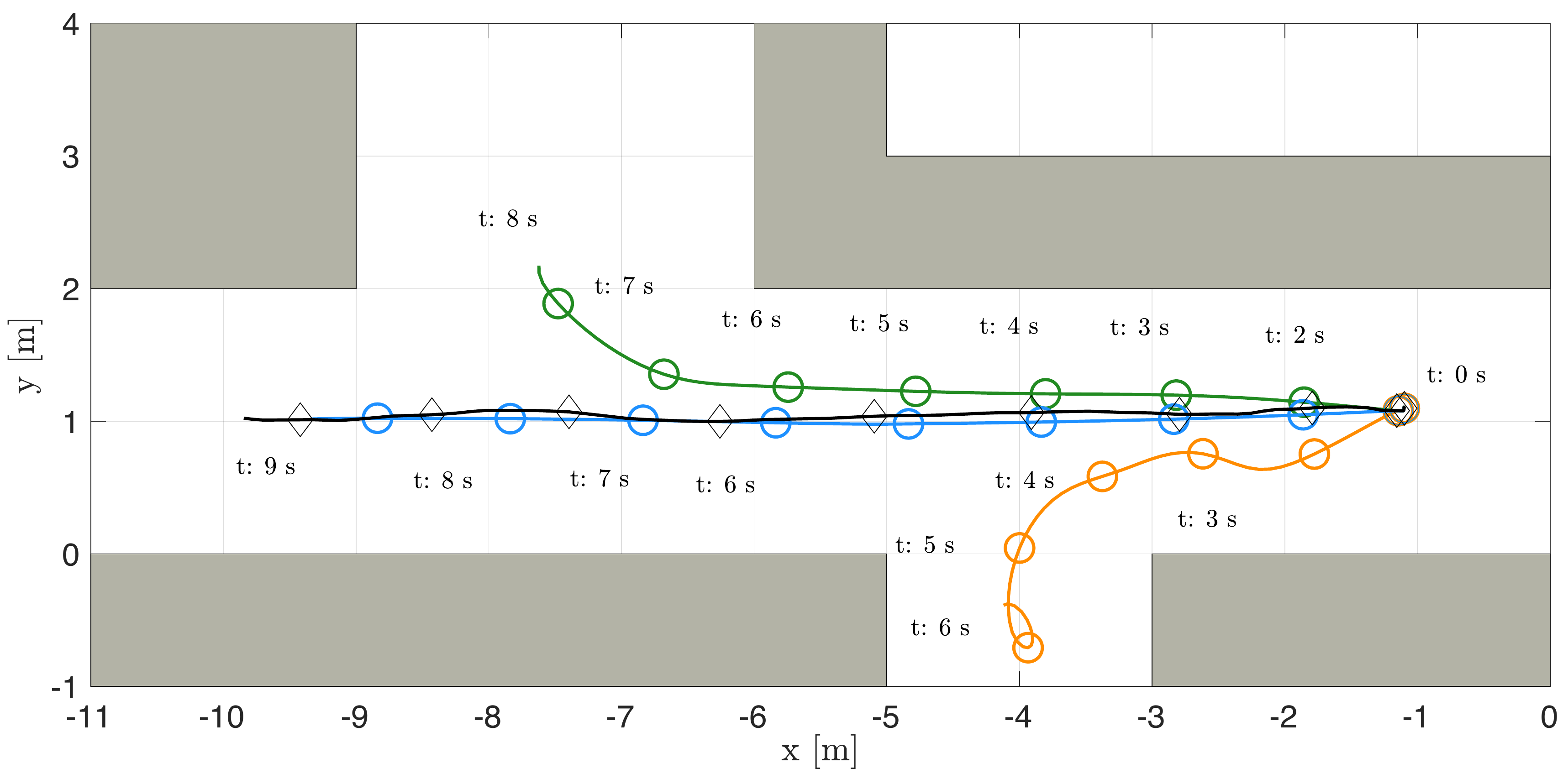} \\
    (a) \\
    \includegraphics[width=0.5\columnwidth]{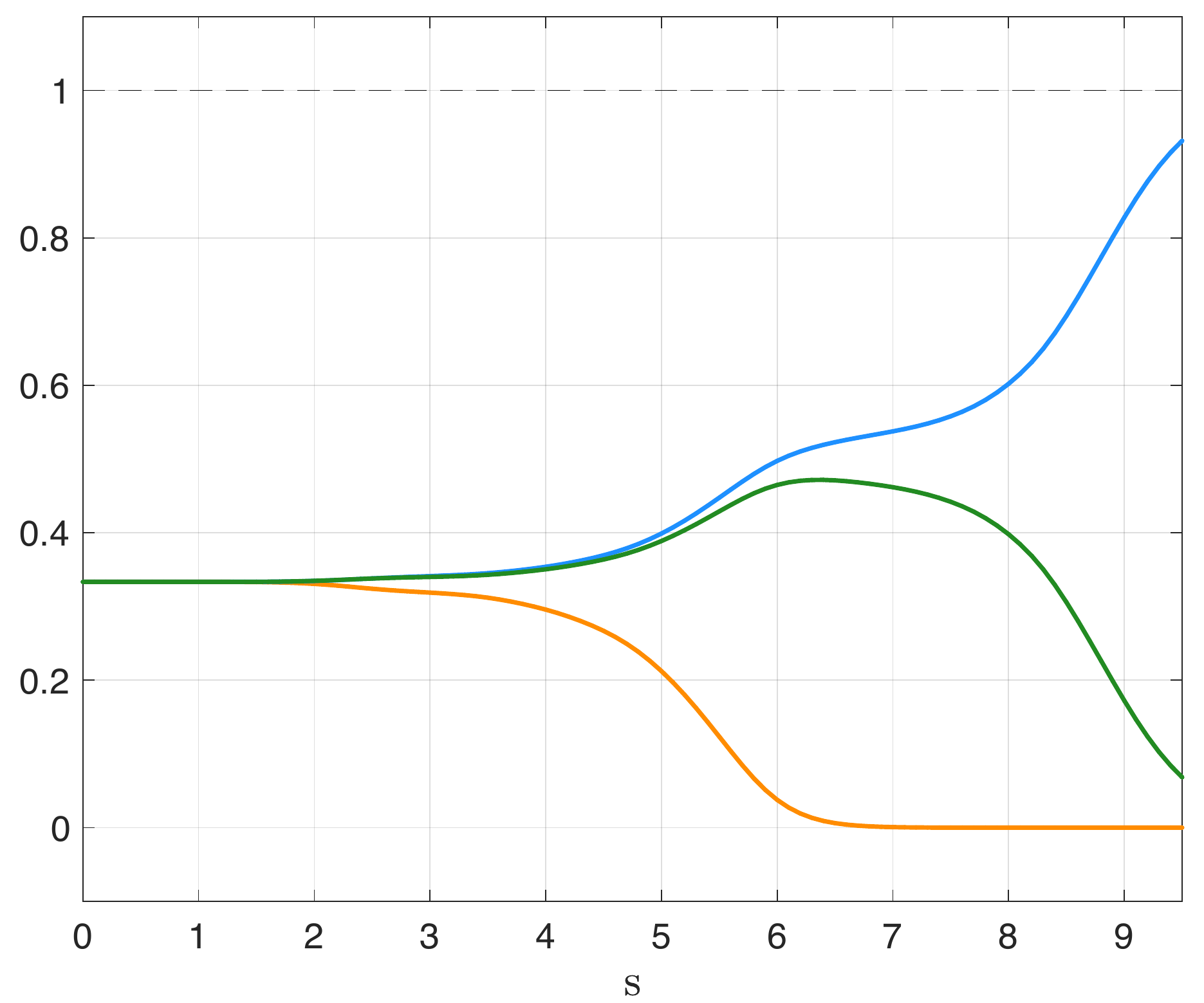} \\
    (b) \\
  \end{tabular}
  \caption{(a) Trajectory predictions compared with the ground truth
    (\textit{black} line). (b) Probabilities of the trajectory
    predictions while the pedestrian moves towards the \textit{blue}
    goal.}
  \label{fig:povo_classify}
\end{figure}

The final evaluation of {\em Net2} with the multi-goal strategy is
carried out through actual experiments in our department at the
University of Trento. In particular, we record the data in a portion
of an hallway with multiple exits. Data were collected using a LIDAR
with a view of $360^\circ$ and maximum measuring distance up to $6$~m
running at $20$ frames per second. The laser scanner was placed about
$80$~cm from the ground at the center of the scene, in order to
optimally see the two sides of the hallway. The measurements points
provided by the sensor were used to both extract the walls information
(that is, the static points between subsequent frames) and the
pedestrian positions. Our acquisition algorithm was used to extract
points belonging to the person waist, and clustered them into a single
planar position. In the recorded set, depicted in
Fig.~\ref{fig:povo}-a, the pedestrian could go to three different
targets, i.e. one directly to the left, one to the right and one right
at the end of the hallway (see Fig.~\ref{fig:povo}-b).
In Fig.~\ref{fig:povo_classify}-a we report the classification result
for trajectory of the \textit{blue} waypoint.
After observing $1$~second of the real trajectory and knowing the
three possible waypoints, we foresee the trajectories in open loop
with the network and the multi-goal strategy. The waypoint on the left
exit is discarded as canditate goal after about $4$~seconds, while the
confidence of the two remaining waypoints remains almost the same,
until the correct goal is found after about $6$seconds, before the
pedestrian oversteps the next exit (see
Fig.~\ref{fig:povo_classify}-b).  It is worthwhile to note that just
one second of observed trajectory is needed and that useful
predictions can be derived by using as a motion strategy a simple
dynamic model such as the SFM. Even though the SFM is not suitable for
motion planning applications (mainly due to its inability to manage
nonlinearities), the synergy with a NN leads to accurate forecast the
local motion avoiding the choice of parameters.

\section{Conclusion}
\label{sec:conclusion}

In this paper, we have shown a novel technique for predicting human
motion. Our idea is based on the combination of a neural network with
a famous physics inspired dynamic model, the SFM. In the combination,
each of the two approaches emphasises its own strengths and
compensates for the weakness of the other. Specifically, the SFM
brings a structure to the NN, reducing its complexity and the number
of samples needed for the training. Furthermore, the NN predictions
become explainable and physically interpretable.  On the other hand,
the NN expresses its full power in terms of flexibility, and of its
ability to learn the complex parameter set of the SFM, which would be
very difficult to estimate in real-time by conventional means for the
strong non linearities of the model.  Our simulations and experiments
reveal the full potential of the marriage between the two worlds of
physics inspired models and neural networks.

Many important points remain open and will attract our efforts in the
near future. First, we aim to establish a full comparison between the
performance of our structure NN and a standard DNN over a number of
realistic use cases. Second, we plan to develop NN embedding different
models which are potentially more realistic than the SFM,
first and foremost the HSFM~\cite{farina2017walking} of the
PHSFM~\cite{antonucci2018towards}. Third, we plan to develop motion
planning algorithms designed to make the best use of learning
in predicting the human motion.

\bibliographystyle{IEEEtran}
\bibliography{SfmLearning}

\end{document}